\documentclass[letterpaper, 10 pt, conference]{ieeeconf}  % Comment this line out if you need a4paper

\IEEEoverridecommandlockouts                              % This command is only needed if 
\usepackage{amsfonts}
\usepackage{multirow}
\usepackage{multicol}
\usepackage{caption}
\usepackage{cite}

\usepackage{todonotes}

\title{\LARGE \bf
Planning with Learned Subgoals Selected by Temporal Information
}

\author{Xi Huang$^1$, Gergely S\'{o}ti$^{1,2}$, Christoph Ledermann$^1$, Björn Hein$^{1,2}$, and Torsten Kröger$^1$% <-this % stops a space
\thanks{The authors are with $^1$Institute for Anthropomatics and Robotics, Karlsruhe Institute of Technology, 76131 Karlsruhe, Germany, and 
$^2$Robotics and Autonomous Systems, Institute of Applied Research, Karlsruhe University of Applied Sciences, 76133 Karlsruhe, Germany\newline{\tt\small x.huang@kit.edu}}%
}

\begin{document}
\maketitle
\thispagestyle{empty}
\pagestyle{empty}

%%%%%%%%%%%%%%%%%%%%%%%%%%%%%%%%%%%%%%%%%%%%%%%%%%%%%%%%%%%%%%%%%%%%%%%%%%%%%%%%
\begin{abstract}
Path planning in a changing environment is a challenging task in robotics, as moving objects impose time-dependent constraints. Recent planning methods primarily focus on the spatial aspects, lacking the capability to directly incorporate time constraints. In this paper, we propose a method that leverages a generative model to decompose a complex planning problem into small manageable ones by incrementally generating subgoals given the current planning context. Then, we take into account the temporal information and use learned time estimators based on different statistic distributions to examine and select the generated subgoal candidates. Experiments show that planning from the current robot state to the selected subgoal can satisfy the given time-dependent constraints while being goal-oriented.
\end{abstract}

%%%%%%%%%%%%%%%%%%%%%%%%%%%%%%%%%%%%%%%%%%%%%%%%%%%%%%%%%%%%%%%%%%%%%%%%%%%%%%%%
\section{Introduction}
Modern robotic applications aim to place the robots into a collaborative environment rather than in a confined workstation, e.g. encapsulated by a safety fence. Such collaborative workspaces are open, allowing humans, robots, and objects to enter them freely. From the perspective of a robot, task-unrelated agents entering its workspace are obstacles that need to be detected and avoided while executing its own task. This avoidance problem is an extension of the classical collision avoidance for static environments, where part of the workspace is blocked by non-moving obstacles. 
%This changes with moving obstacles (dynamic environment): 
Previously, being blocked or not was a static spatial property, based on the spatial occupation and position of the obstacles. Now, in a collaborative environment, this property becomes time-dependent, highly increasing the complexity of the problem.

In this work, we propose to simplify the problem posed by changing environments with the assumption that a changing environment is static over a small period of time, where the changes are so small that they do not affect path planning. 
Hence, an overall path can be planned and executed step by step, always assuming that the current environment state is static. For static environments, a variety of sampling-based planning approaches have been proven to be efficient. Among recent methods, some try to estimate meaningful spatial distributions for sampling \cite{chamzas2021learning, lee2022adaptive, zhang2018learning, kumar2019lego, ichter2018learning, ichter2019learned}, some propose to approximate the cost-to-go function \cite{huh2020Learning}, while some neural planners imitate the behavior of the optimal planners and plan a path \cite{qureshi2020motion} or output an action directly \cite{fishman2023motion}.
%Although they differ in their approach to sampling or planning, they all rely on the spatial domain alone to perform their computations. 
%However, our general approach assumes that the environment is fixed for only a small, finite amount of time. Planning has to start during this time window, making it time-dependent. None of the existing approaches allow us to incorporate this time dependence directly.
Despite the diversity of these methods, all these methods rely on the spatial domain alone to perform their computations. However, our approach has an assumption that the environment is fixed for only a small time window, imposing a time constraint for the planning. None of the approaches above allow us to incorporate this time constraints directly.

Therefore, in our work, we aim to integrate temporal information into learning to make the algorithm aware of our time budget while solving a planning problem. In this context, temporal information refers to the planning time used to plan from the current robot state to intermediate subgoals. %Together with this temporal information, the recent methods, which only learn from the spatial information, develop a new opportunity to predict the important samples in scenarios where there is a constraint regarding planning time, such as changing environments. This preserves the advantage of sampling-based motion planners, i.e., a collision-free path without getting into a local minimum. 
Together with this temporal information, the recent methods, which only learn from the spatial information, can be extended and applied to scenarios where there are constraints regarding the planning time, such as changing environments. 

\begin{figure}
    \centering
    \includegraphics[width=0.46\textwidth]{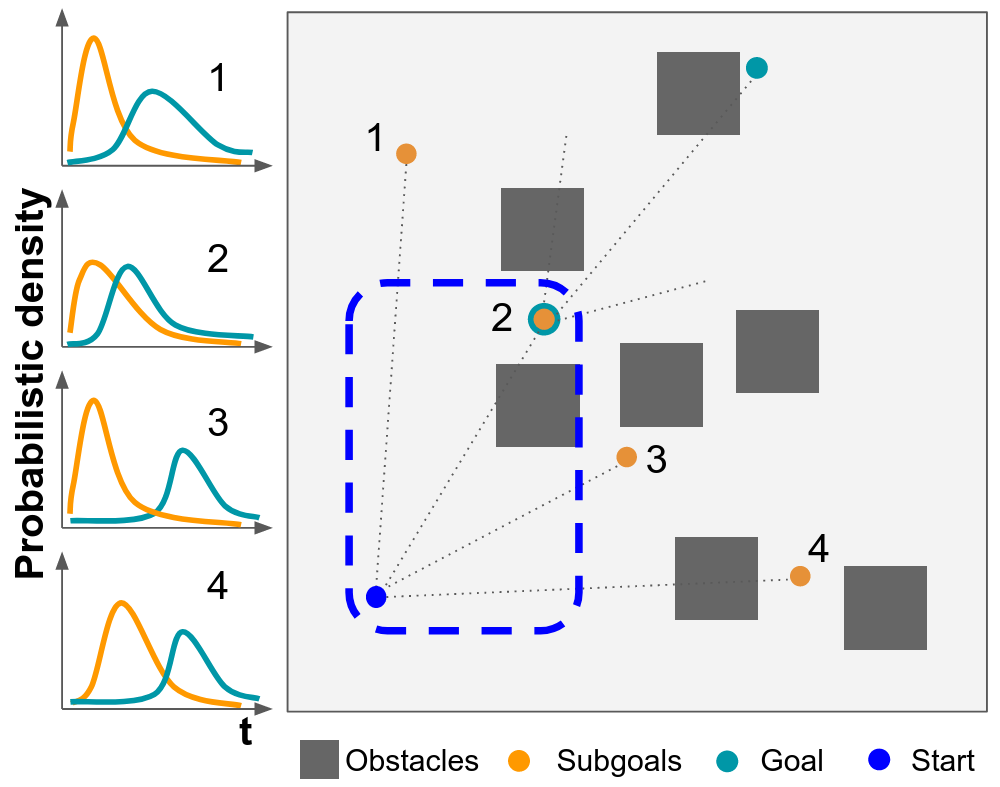}
    \caption{Planning with learned subgoals: with the estimated probabilistic distribution of planning time from the start to subgoal candidates (orange), and from the goal to subgoal candidates (turquoise), candidate 2 is selected. Planning ranges are adapted to the blue dashed bounding box after the selection for planning efficiency.}
    \label{fig:first_page}
\end{figure}

Combining the spatial and temporal information, we propose Planning with Learned Subgoals (PLS), which outputs subgoals to incrementally lead the robot to the final goal. 
Unlike the methods that use a learned spatial distribution to bias the sampling process, the subgoals offered by our approach are more like milestones. % that lead the robot in a straight-forward direction.
These milestones decompose a complex planning problem into small and easily solvable pieces. For small and easy problems, we can further tailor the planning range, in which we sample and plan, 
to achieve better performance. As illustrated in Fig. \ref{fig:first_page}, given a planning problem with obstacles, we generate a batch of candidates for subgoals, marked in orange. Then, based on the estimation of the planning time from the start to the candidates and from candidates to the goal, we select the most suitable candidate with index 2 as a subgoal according to the metrics introduced in \ref{subsec:subgoal_select}. After the selection, the range of the planning problem presented in \ref{subsec:range} shrinks to the blue dashed box for better planning performance. 
Experiments show that the integration of the temporal information makes the planning algorithm significantly more effective and can be applied to reactive planning scenarios where the planner has limited time to find a solution. 
%In the experiments, our method generates sub-goals that can be planned with 0.016 seconds on average while the baseline method needs 1.15 seconds on average. Accumulating time consumption between the intermediate subgoals toward the goal of the planning problem, our method achieved a result of 0.149 seconds on average, which also shows advantages compared to baselines.
Our contributions are mainly:
\begin{itemize}
    \item Leveraging a generative model to predict spatial subgoals that decompose a complex planning problem. 
    \item Using learned time estimators to capture the temporal information regarding the planning time given the planning problem.
    \item Designing two metrics based on the time estimation to select suitable and goal-directed subgoal candidates. 
    \item Conducting an ablation study to compare the performance based on different assumptions of the distribution for time estimators.
\end{itemize}

% The rest of the paper is constructed as follows: we review the related work in \ref{sec:related_work} and outline the difference between this approach to them. 

\section{Related Work}
\label{sec:related_work}
Recent approaches combining machine learning and sampling-based motion planning fall roughly into two categories, i.e., learning the heuristics or distributions and neural planners. We review them in the following and highlight the differences between these methods. 

\subsection{Learning heuristics}
Heuristics are crucial for speeding up the planning process in the sampling-based methods. They serve as an oracle to guide the search or sampling \cite{gammell2015batch, dellin2016unifying, strub2022Adaptively, huang2022hiro}. Recent approaches attempt to learn various kinds of heuristics to achieve faster planning. Planning delay is featured as a heuristic in \cite{kaur2021speeding} to keep the planner away from the local minimum, which usually has a high delay. Uncertainty is used as a heuristic to guide the planner to explore in \cite{hou2020posterior} by updating probabilistic models using maximum posterior every time after gaining new information. Given a planning problem, \cite{huh2020Learning} learns the cost function as heuristics and uses the gradient of the cost function to find a path between the start and the goal.  

\subsection{Learning distributions}
While learning heuristics aims to estimate a value given a sample or region and use this value to prioritize the planning or searching, learning distribution directly predicts the regions likely to solve the problem. Using a conditional variational autoencoder (CVAE) to capture latent representations, \cite{ichter2018learning} learns a sampling distribution of the optimal path given the planning requests. \cite{ichter2019learned} applies the connectivity in graph theory to identify the critical configurations in a roadmap. These critical configurations improve the capability of a roadmap to solve narrow passage problems. LEGO \cite{lee2020magic} proposes a method to collect the bottleneck configurations. Then, they leverage a CVAE as a generative model to generate bottleneck configurations during the inference. These bottleneck configurations serve the same purpose as in \cite{ichter2019learned}.
%[Although CVAE is a good tool to generate samples, the loss must be carefully designed to balance the reconstruction and multi-modality.] 
The methods above learn some predefined features and metrics using a supervised manner. The authors of \cite{lee2022adaptive} notice the gap between sampling and the downstream planner performance. They propose a generator-critic architecture, similar to actor-critic in reinforcement learning. The critic serves as a proxy to quantify whether the samples contribute to finding a solution. This links the sampling distribution directly to the planning performance, e.g., planning time and path quality. Learning distributions from experience can also be found in \cite{chamzas2021learning} \cite{chamzas2022learning}.
While the distribution is explicitly learned and deployed during the sampling phase, \cite{zhang2018learning} learns an agent to reject implausible samples. Implicitly, the agent learned the knowledge of the plausible region. Rather than learning the distribution from collected data, \cite{lambert2022Stein} applies the Stein variational inference to shifting samples toward the free space by optimizing a posteriori. 
\begin{figure}
    \centering
    \includegraphics[width=0.48\textwidth]{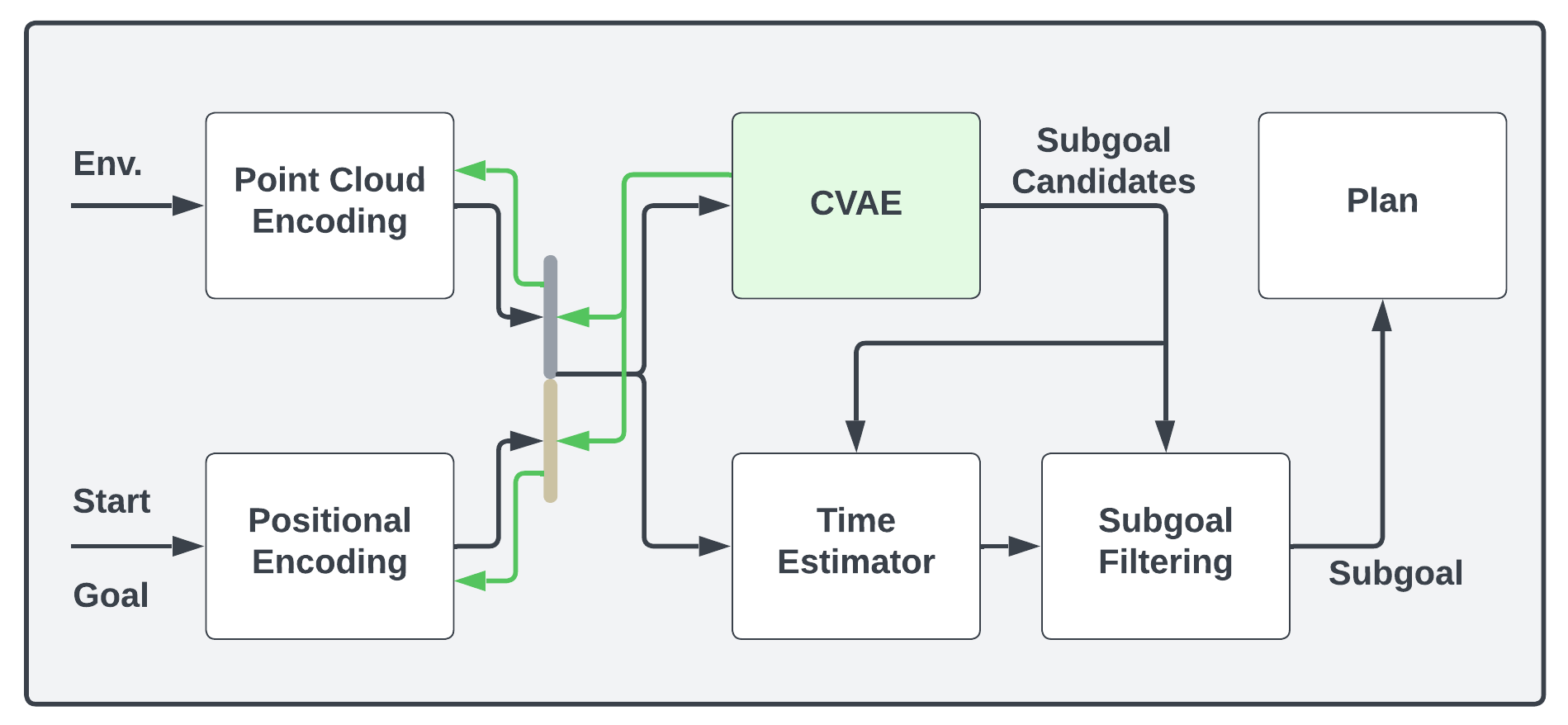}
    \caption{The PLS architecture. The green arrows show the direction of gradient descent. During training, the backpropagation of the time estimator does not reach the encoding blocks.}
    \label{fig:architecture}
\end{figure}
\subsection{Neural planners}
The methods above learn to predict the distribution of the sampling or heuristics for planning and eventually use a traditional planner to find a solution. Different from that, neural planners try to mimic the behavior of traditional planners and learn directly to plan. 
Motion planning network \cite{qureshi2020motion} imitates the behavior of the RRT* planner and outputs a complete path by planning in a bi-directional fashion. It utilizes offline learning and active learning to reinforce the planning performance. Motion policy networks \cite{fishman2023motion} mimic the behavior of diverse planners in both task and configuration space, including AIT* \cite{strub2022Adaptively} and Geometric Fabrics \cite{xie2020geometric}. Instead of learning to capture the whole path, it learns to generate the delta, i.e. the joint velocity given the current planning problem.

Our approach goes beyond existing methods that focus solely on spatial information by incorporating temporal data into the planning process. We train models to predict the time required for planning using a normal or log-normal distribution. These predictions help us identify the most useful samples produced by the generative model, ensuring that the planning process stays within a specific time frame while still goal-oriented. As the planning time is strongly correlative to the number of collision checks, estimating the planning time is a new and closer way to link the samples to the planning performance. 

\section{Method}
Given a planning problem, including the start and goal joint configuration, and the information of the surrounding environment, the objective of this work is to generate joint configurations of a robot manipulator as subgoals that not only lead the robot towards the final goal configuration but also can be planned within a given time constraints. We address this problem by using a generative model to output a batch of candidates for an intermediate subgoal instead of planning a complete path. Then, we employ the temporal estimation regarding the planning time in two different aspects to select the proper one as the subgoal. Fig. \ref{fig:architecture} shows the architecture of PLS.

\subsection{Dataset}
To integrate both temporal and spatial factors, we have assembled a dataset containing over one million planning requests using OMPL\cite{sucan2012the-open-motion-planning-library}. These requests were planned using a UR10e robot in various environments with at least one feasible solution. Each request provides a unique, collision-free starting and ending configuration that lies within $[-2\pi, 2\pi]$ for all joints. Each data entry in our dataset contains information about the environment, the start and final configurations, the intermediate waypoints between them, and the time needed for planning from the starting point to each waypoint. Multiple runs were conducted to obtain a comprehensive distribution of planning times.

For the generation of spatial waypoints connecting the start and goal configurations, we employ the AIT* optimal planner. Subsequently, we compute the time required to traverse from the start configuration to each waypoint. We choose RRTConnect\cite{kuffner2000rrt} over AIT* for the computation of planning times. This choice is informed by the characteristics of AIT* as an optimal planner that continually refines its solutions until the given time budget is over. Our observations indicate that AIT* generally takes a bit more time to return an initial solution compared to RRTConnect, which is consistent with the results reported in the original AIT* paper \cite{strub2022Adaptively}. %This observation is consistent  reported in the original AIT* paper\cite{strub2022Adaptively}. 
% Additionally, AIT* exhibits a broader distribution of planning times. 
\label{method:dataset}
\begin{figure}[!]
    \centering
    \includegraphics[width=0.45\textwidth]{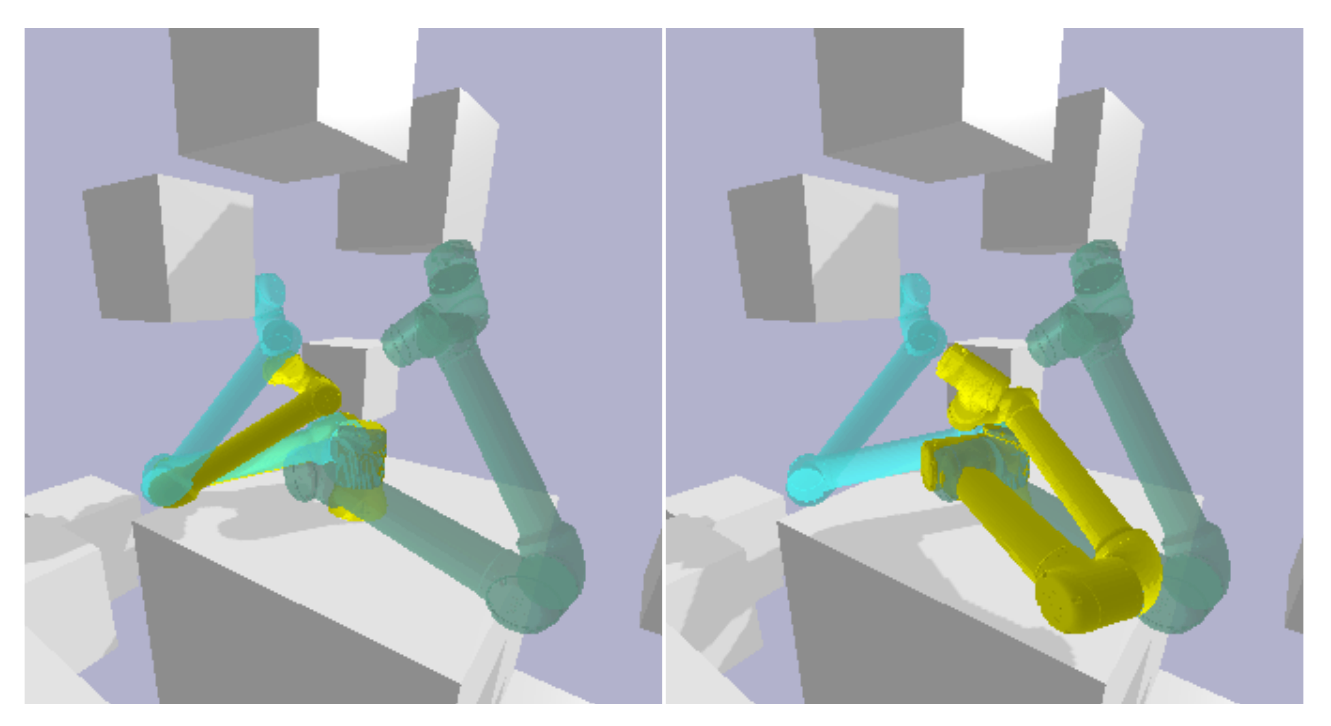}
    \caption{An example of the dataset. The start and goal configurations are marked in cyan and green, respectively. The subgoals, namely ground truths are shown in yellow. }
    \label{fig:enter-label}
\end{figure}

\subsection{Generating Spatial Subgoals}
\label{method:spatial}
Planning to subsequent subgoals offers several benefits compared to directly aiming for the global goal. 
First, the planner typically needs less planning time to return a solution. This efficiency arises from the reduced distance between the starting point and the subgoal. Empirical evidence suggests that planning time tends to grow exponentially with the increase in distance.
The subgoals decompose complex planning problems into pieces and are usually in closer proximity to the current robot state. While the environment changes, paths leading to subgoals in closer proximity are preferable. This is because paths to global goals, which are usually longer, are more vulnerable to becoming invalid due to these changes.
%Generally, the planning range should be large for such complex planning problems.
An additional advantage is that we can accordingly shape and narrow the planning range based on the subgoals, thereby minimizing the search effort. 
This is especially beneficial in spaces with high degrees of freedom.
 % However, these advantages come only with the condition that subgoals are final-goal-oriented, located in the region near the optimal paths, ensuring they progressively guide the robot toward its final goal.
Note that these are advantages only under that condition that the subgoals are plausible and goal-oriented, meaning they can progressively guide the robot toward its final goal.

%Therefore, it is crucial to generate plausible subgoals. 
Previous works \cite{ichter2018learning, kumar2019lego} showcase that a conditional variational autoencoder (CVAE) \cite{sohn2015learning} is capable of capturing the environment representation and generating samples that are similar to the dataset, e.g. generating samples along the optimal path or in the bottleneck region. Similar to them, we employ a CVAE to generate plausible subgoals given the planning problem as conditions. Different from them, we take temporal information into account and train our CVAE in such a way that the model tends to generate subgoals that can meet the time constraints. To achieve this goal, during the training, we only use the waypoints in the dataset mentioned above with a constraint for maximal planning time of 0.05s. Theoretically, this constraint for planning time should be small to enable reactive planning in changing environments. Practically, however, the smaller the range is, the fewer generated data can fit into it and thus training a model properly becomes harder. %Due to this trade-off, the maximal planning time is set to be 0.05s in this study. 

Given a planning problem $\mathbf{c} \in \mathcal{C}$, including the start and goal configuration and the surrounding environment, we aim to generate the subgoals that can lead the robot to reach the goal. The CVAE model samples $\mathbf{z} \in \mathbb{R}^m$ from a $m$-dimensional i.i.d. latent space and produces $n$-dimensional joint configurations $\mathbf{x} \in \mathbb{R}^n$. 
%The mapping from $z$ to $x$ is usually done by a neural network with parameters $\theta$.
The objective is to maximize the probability that the model generates the data that  
are likely to be in the dataset $\mathcal{X}_{sg}$ over the whole $\textbf{z}$-space
\begin{equation}
    p(\mathbf{x}{|}\mathbf{c}) = \int p(\mathbf{x}|\mathbf{z}, \mathbf{c}) p(\mathbf{z}) \mathrm{d} \mathbf{z}.
\end{equation}
%To ensure that $\mathbf{z}$ is a representative latent variable, the distribution $q(\mathbf{z}|\mathbf{x}, \mathbf{c})$ of the latent variable of the dataset $\mathcal{X}_{sg}$ and the $p(\mathbf{z}|\mathbf{x}, \mathbf{c})$ we used to generate the synthetic data should be similar.
As we aim to generate data that are similar to the dataset $\mathcal{X}_{sg}$, the distribution of the latent variables for generating the synthetic data $p(\mathbf{z}|\mathbf{x}, \mathbf{c})$ should be similar to the distribution $q(\mathbf{z}|\mathbf{x}, \mathbf{c})$ represented by the dataset. 
The Kullback-Leibler (KL) divergence between these two distributions is
\begin{equation}
    \mathcal{D}[q(\mathbf{z}|\mathbf{x}, \mathbf{c}) || p(\mathbf{z}|\mathbf{x}, \mathbf{c})] = \mathbb{E}_{\mathbf{z} \sim q}[\log q(\mathbf{z}|\mathbf{x}, \mathbf{c}) - \log p(\mathbf{z}|\mathbf{x}, \mathbf{c})].
\end{equation}
In the following, we use $\mathcal{D}[\cdot]$ to denote the KL divergence. The evidence lower bound (ELBO) of $\log p(\mathbf{x}|\mathbf{c})$ can be written as:
\begin{equation}
    \log p(\mathbf{x}|\mathbf{c}) \geq
    \mathbb{E}_{\mathbf{z} \sim q}[\log p(\mathbf{x}|\mathbf{z}, \mathbf{c})] - \mathcal{D}[q(\mathbf{z}|\mathbf{x}, \mathbf{c})||p(\mathbf{z}|\mathbf{x}, \mathbf{c})].
\end{equation}
While we maximize the evidence lower bound, the $\log p(\mathbf{x}|\mathbf{c})$ will increase accordingly. We suggest the reader refer to \cite{doersch2016tutorial} for further information regarding CVAE. 
The first term of the right-hand side characterizes the likelihood of a data point $\mathbf{x}$ given the $\mathbf{z}$ and $\mathbf{c}$. Given the assumption that $p(\mathbf{x}|\mathbf{z}, \mathbf{c}) \sim \mathcal{N}(f(\mathbf{z},\mathbf{c}),\mathbf{\Sigma})$, where $\mathbf{\Sigma}$ denotes diagonal covariant matrices and $f(\mathbf{z},\mathbf{c})$ is the generated sample,  we have 
\begin{equation} \label{eq:log_likelihood}
    \log p(\mathbf{x}|\mathbf{z},\mathbf{c}) \propto - ||\mathbf{x}-f(\mathbf{z},\mathbf{c})||^2_2
\end{equation}
The assumption $p(\mathbf{x}|\mathbf{z},\mathbf{c}) \sim \mathcal{N}(f(z,\mathbf{c}),\mathbf{\Sigma})$ has been used for various tasks and we consider it valid within the scope of this paper. We modify Eq. \ref{eq:log_likelihood} to be 
\begin{equation} \label{eq:g_log_likelihood}
    \log p(\mathbf{x}|\mathbf{z},\mathbf{c}) \propto - ||g(\mathbf{x})-g(f(\mathbf{z},\mathbf{c})||^2_2, 
\end{equation}
with a non-linear function capturing both the forward kinematics $g_{FK}(\cdot)$ and joint positions $g_{joint}(\cdot)$ of a robot using a weighting factor $\alpha$
\begin{equation}
    g(\mathbf{x}) = \alpha g_{FK}(\mathbf{x}) + (1-\alpha) g_{joint}(\mathbf{x}).
\end{equation}
Our model directly outputs joint configurations. However, predicting wrong values at the first joint and at the end-effector can generally lead to significantly different effects. The changes at the first joint usually will result in a larger placement of the robot. Therefore, the $g_{FK}$ is a weighting function that reflects the kinematics of the robot. However, for a robot with a range of $[-2\pi, 2\pi]$ at every joint, the forward kinematics can be the same although the joint configurations are different. To address this problem, we introduce the positional encoding with $l$ levels
\begin{equation}
    g_{joint}(\mathbf{x}) = [\mathbf{x}, \cos{\mathbf{\mathbf{x}}}, \sin{\mathbf{x}}, ..., \cos{(2^{l}\mathbf{x})}, \sin{(2^{l}\mathbf{x})}].
\end{equation}
Positional encoding is used in recent methods like Neural Radiance Field (NeRF) \cite{mildenhall2021Nerf} to capture the high-frequency part of the data. In our case, positional encoding provides the model with more information to understand the planning problem. 
By assuming the prior of the latent space to be  $p(\mathbf{z}|\mathbf{x},\mathbf{c}) \sim \mathcal{N}(\mathbf{0}, \mathbf{I})$, the final loss of the CVAE becomes
\begin{equation}
    \mathcal{J} = ||g(\mathbf{x})-g(f(\mathbf{z},\mathbf{c}))||^2_2 + \beta \mathcal{D}[q(\mathbf{z}|\mathbf{x}, \mathbf{c})||\mathcal{N}(\mathbf{0}, \mathbf{I})],
\end{equation}
with a weighting factor $\beta$\cite{higgins2016beta}. Two neural networks are used to capture approximate the function $q(\mathbf{z}|\mathbf{x}, \mathbf{c})$ and $f(\mathbf{z}, \mathbf{c})$, respectively. Using the reparametrization trick \cite{kingma2013auto}, we apply stochastic gradient descent to train the networks.

%In the training phase, the model should mimic the subgoal distribution $q(x|c)$, where $x \in \mathcal{X}_{sg}$ is an instance of the subgoal distribution $\mathcal{X}_{sg} \subset \mathcal{X}_{free}$. The distribution $p(x|c)$ learned by the model should be close to the data distribution $q(x|c)$.

%where the $z$ is a latent representation of the data. In the generation phase, the model samples values from $z$ and produces samples likely to be in $\mathcal{X}_{sg}$. 

\subsection{Learning Temporal Distributions}
\begin{figure}[!]
    \includegraphics[width=0.48\textwidth]{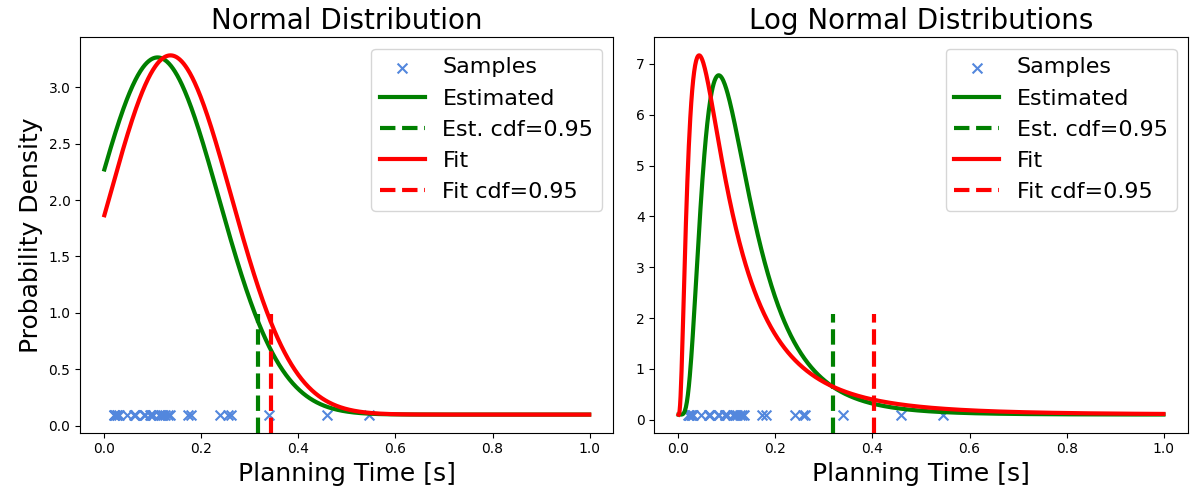}
    \caption{Samples of planning time described by normal and log-normal distributions. Red: statically determined. Green: estimated by a model.}
    \label{fig:ditribution}
\end{figure}
While using a generative model to learn subgoals, it is desired that the model can provide diverse and multi-modal subgoals, which can solve the problem differently. On the other hand, diversity can lead to a problem because the planning time toward the subgoal may not fit the time constraints. To address this problem, we employ a time estimator to select the proper subgoals regarding the constraints. The time estimator takes the environment, the start configuration, and the subgoal as input and outputs a distribution of the planning time. This temporal distribution serves as a proxy of the planning effort a planner needed for the problem and it is strongly correlative to the number of collision checks.

After analyzing the data, we consider normal and log-normal distribution. Log-normal distribution describes the data whose logarithm is normally distributed. In other words, given the normally distributed data $X$, log-normal distribution describes the data $Y = e^X$. The data from a log-normal distribution is always positive and skewed. As shown in Fig \ref{fig:ditribution}, the data has a long tail in the positive direction.
%Some literature suggests the inverse Gaussian instead. Due to the PyTorch support, we would not consider it in our work.

As mentioned before, the objective of learning temporal distribution is to verify if the predicted subgoal can be planned within the time constraints. Intuitively, the time distribution depends on the complexity of a planning problem, which is defined by the start-goal query and the environment. Therefore, we use the same encoded latent representations in \ref{method:spatial} as a starting point to learn the time distribution. A multi-layer perceptron (MLP) takes this encoded representation as input and outputs the parameters $\hat \theta$ of a 1-D distribution. To match the planning time distribution for each planning instance $i$ in the dataset depicted \ref{method:dataset}, we form the loss function based on negative log-likelihood
\begin{equation}
    J_i(T_i, \hat \theta_i) = -\frac{1}{N} \sum_{j=0}^N \mathcal{L} (t_{i,j} | \hat{\theta} ) + w||\theta_i - \hat{\theta}_i||_2^2,
\end{equation}
where $T_i \in \mathbb{R}$ is a random variable and $t_{i, j} \in T_i$ is a planning time collected by the dataset. In addition to the negative log-likelihood, we add a mean square error (MSE) loss between the parameter $\theta$ empirically derived from the data $t_j$ and the output of the time estimator $\hat{\theta}$, weighted by $w$. As shown in Fig. \ref{fig:architecture}, the gradient of the time estimator does not backpropagate to the encoding blocks. We employ this time estimator to inform the selection of subgoals, described in the following in \ref{subsec:subgoal_select}.

\subsection{Subgoal Selection based on Temporal Estimation}
\label{subsec:subgoal_select}
We designed two selection metrics based on the temporal estimation to select the subgoals that 
are likely to be planned within the time constraints, and that are goal-oriented, respectively.
Selecting goal-oriented subgoals is important, because, in an extreme case, any model that copies the start configuration from the condition and uses it as output, would meet the time constraints. 

% The first metric, also the essential one, is the collision metric. Although it is not often that the model generates samples colliding with the environment, we set this metric as the first filter of the selection. The filtered configuration will be further selected according to two temporal metrics. 

% These two metrics are based on the time estimation model mentioned above. The first temporal metric, the start-to-sample, characterizes the distribution of planning time from the start configuration to the generated samples. With this metric, we can estimate how likely the planning time from the start and generated samples will exceed the user-defined time constraints. The second temporal metric, the goal-to-sample metric, ranks the estimated time distribution from the goal to the sample. This metric depicts if the sample is goal-oriented.

\subsubsection{Start-to-sample metric}
The start-to-sample metric estimates how likely the planning time from the start and generated subgoal candidates will exceed the user-defined time constraints. With the estimated parameters $\hat \theta$, we can analytically determine the cumulative density function (CDF) $c(\cdot)$ of the estimated distribution. As shown in Fig. \ref{fig:ditribution}, the vertical dashed lines indicate the position $t_{95}$ where the $c(t_{95}) = 0.95$, meaning that 95\% of the values drawn from this distribution are supposed to be smaller than $t_{95}$. We see $t_{95}$ as a threshold to classify if the samples can meet the time constraints. Depending on the needs, one can set this confidence level to a higher value.

%Comparing the desired time budget $t_{d}$ to $t_{95}$, we can filter out the samples estimated to be plannable. If none of the samples fits the condition $t_{95} \leq t_d $, we may either choose the sample with the lowest $t_{95}$ value or choose the one with the highest $c{t_d}$. 

\subsubsection{Goal-to-sample metric}
While the start-to-sample metric classifies if the samples can be planned within the time constraint, it does not quantify how goal-oriented the samples are. In search algorithms, the cost-to-go heuristics depict the estimated cost of reaching the goal and are used to prioritize the exploration. These heuristics usually are spatial distances such as the Manhattan distance. Similar to the cost to go, we can estimate the time distribution from the goal to the subgoal candidates using the same time estimator model. We further use the $t_{95}$ to describe how are these candidates close to the goal. % Other characteristics such as mean or mode can lead to similar results in ranking the samples. 
%In addition, the goal-to-sample metric can be used as the trigger if the sample is close to the goal.

These two metrics can be applied to the selection separately or together. In the experiment described in \ref{exp:1}, we introduce two selection criteria, named best-effort and goal-oriented, respectively. The best-effort selection first computes the $t_{95}$ value of all subgoal candidates and randomly chooses a candidate from the ones with $t_{95} \leq t_d$, where $t_d$ denotes that the time constraints for planning. If no candidates meet the constraints, it selects the one with the smallest $t_{95}$. The goal-oriented selection builds on the best-effort selection. Instead of randomly choosing the qualified candidates with $t_{95} \leq t_d$, it ranks the samples by the goal-to-sample metric and selects the one with the smallest value, i.e. the most goal-oriented one. 

\subsection{Planning range shaping} \label{subsec:range}
A larger planning range usually means a larger search space and therefore a longer planning time for sampling-based motion planners. For some complex planning problems, we have to keep the planning range large. As the subgoals break down the complex problems into pieces, we are able to shape the range accordingly for every small planning problem.
This concentrates the samples in the region where the planner can find a solution. A similar concept can be found in batch-informed trees (BIT) \cite{gammell2015batch}. The difference is that the methods in the BIT family draw the bounds formed by high-dimensional ellipsoids after an initial solution is found. In our case, we directly set the planning range based on the planning problems.

For an $n$-DoF robot arm, we set the lower bounds $\mathbf{b}_l \in \mathbb{R}^n$ and upper bounds $\mathbf{b}_u \in \mathbb{R}^n$ for joints depending on the mobility of each joint and the planning problem. We constrain the search space with greater paddings for the joints that can produce large motion, usually the ones close to the robot base, depending on the robot's kinematics. 

\section{Experiments and Results}
\subsection{Quantitative Evaluation}
\label{exp:1}
% We evaluate our method by planning in a large dataset with more than 10,000 diverse planning scenarios. There is at least one feasible path in all of these scenarios. We investigate the performance of the CVAEs with three aspects. First, we investigate if the samples generated by the CVAEs can be planned within the time constraints. Second, we investigate if the subgoals generated by CVAEs can lead to the global goal. Third, we compare the path formed by a sequence of subgoals to the baselines. 

% First, we use the RRTConnect and the vanilla CVAE as baselines. We plan the scenario with RRTConnect from the start to the global goal, record time distribution, and use it as the baseline. For the baseline of vanilla CVAE, we use it to generate subgoals, plan from the start or the last subgoal to the new generated sub-goal, and finally record the time distribution. We follow the same procedure for other CAVEs. As listed in Table [??], we evaluate the performance of the CVAEs with selection with time constraints, CVAE jointly trained with time estimator, with and without selection.

% Second, we roll out the procedure until it reaches the goal or the number of rollouts exceeds the pre-defined number of trials. We compare the success rate of the CVAEs models and the accumulated planning time. Note that the success rate of RRTConnect is certainly 100\% since we use it to generate the dataset.

% As shown in the Table [??], we compare the path length for the successful cases to the baselines, in this case, the RRTConnect and AIT*.

We verify and evaluate the effectiveness of our proposed method within two steps. First, we verify that the model can generate samples that can be planned within a short period of time. As an ablation study, we compare the results of the generated samples with and without the adaptive planning range. The difference between the selctions based on normal and log-normal distribution is shown as well.
Secondly, we examine the capability of CVAE-generated subgoals in guiding the robot toward the final goal. In this part, we contrast the planning time as well as the quality of the paths formed by a sequence of subgoals against established baselines. %The experiment goes through over 10k unique planning problems. Each problem guarantees the presence of at least one feasible solution path.

\begin{table}[b]
    \centering
    \begin{tabular}{|l|l|l|l|l|l|}
    \hline
    & \multicolumn{3}{c|}{Distribution [\%]} & \multicolumn{2}{c|}{Time [s]} \\ \cline{2-6}
    & 0.05 &  0.1  & 0.2 & Mean & Std. \\ \hline \hline
    RRTConnect  & 0 & 3.8 & 12.9 & 1.153 & 1.86 \\
    Random & 65.4 & 74.9 & 83.2 & 0.143 & 0.328 \\
    B-L-S  & 85.1 & 90.0 & 94.1 & 0.056 & 0.203 \\
    B-N-S  & \textbf{89.2} & \textbf{93.8} & \textbf{96.7} & \textbf{0.035} & \textbf{0.137} \\
    B-L  & 67.0 & 73.6 & 79.9 & 0.198 & 0.434 \\
    B-N  & 71.3 & 78.9 & 86.2 & 0.122 & 0.211 \\
    \hline
    \end{tabular}
    \caption{\label{tab:exp1} Planning to a subgoal. The letter ”B” denotes the best-effort selection. The letter ”L” and ”N” indicates the selection using the log-normal and the normal distributions, respectively, while ”S” signifies the planning range shaping.}
    \begin{tabular}{|l|l|l|l|l|l|l|l|}
    \hline
    & Succ. & \multicolumn{3}{c|}{Distribution [\%]} & \multicolumn{2}{c|}{Time [s]} & Length \\ \cline{3-7}
    & [\%] & 0.05 &  0.1  & 0.2 & Path & Subgoal & [rad]\\ \hline \hline
    Baseline & \textbf{100} & 0 & 3 & 11 & 1.153 & - & \textbf{17.65} \\
    Random & 61.7 & 82.6 & 87.4 & 91.9 & 0.517 & 0.061 & 50.67 \\
    B-L-S & 81.5 & 90.1 & 93.2 & 95.9 & 0.208 & 0.029 & 25.79 \\
    B-N-S & 69.0 & \textbf{95.2} & \textbf{97.0} & \textbf{98.4} & \textbf{0.149} & \textbf{0.016} & 23.55 \\
    G-L-S & 85.2 & 89.3 & 93.3 & 96.2 & 0.172 & 0.030 & 25.51 \\
    G-N-S & 82.3 & 88.5 & 92.2 & 96.2 & 0.214 & 0.039 & 25.56 \\
    \hline
    \end{tabular}
    \caption{\label{tab:exp2} Planning to fianl goals. For the baseline statistics, we list the path length of the optimal planner AIT* with a planning budget of five seconds and the planning time of RRTConnect for the same reason mentioned in \ref{method:dataset}.}
\end{table}

For the first aspect, our focus is to verify the capability of the model and outline the contribution of integrating the subgoal selection based on the time estimation and the planning range shaping, respectively. The planning problems used for the experiment are first planned with the RRTConnect planner for 30 runs. All runs of these planning problems result in a planning time of at least 0.05s. Therefore, we can see that RRTConnect has 0\% in the case of less than 0.05s.
As summarized in Table \ref{tab:exp1}, we evaluate the CVAE model using random selection and four other variants. These variants differ from each other based on their assumption of distributions (L or N) and the inclusion or exclusion of planning range shaping (S). In the random selection, we randomly select a candidate from the batch generated by the CVAE model without selection and use it as the subgoal. Each subgoal is planned using RRTConnect with 30 runs as well.
For the four variants, instead of the random selection, we use the best-effort selection mentioned in \ref{subsec:subgoal_select} without considering whether the candidates are goal-oriented, marked as "B" in Table \ref{tab:exp1}. The desired time bound $t_d$ is set to be 0.05s. 
%If there is more than one candidate fulfilling the condition $t_{95} \leq t_d$, the best-effort selection randomly chooses one from them, otherwise, the one with the smallest $t_{95}$ will be chosen. 
%We denoted the best-effort selection with the letter "B" in Table \ref{tab:exp1}. The letter "L" indicates the selection using log-normal distributions and "N" indicates the normal distributions, while "S" signifies the experiment with planning range shaping.
Three key values are defined to capture the distribution of the planning time, namely 0.05, 0.1, and 0.2. For example, the column of 0.05 shows the percentage of the planning time $t \leq 0.05s$. The mean and standard deviation of the planning time are also listed.
Compared to the random selection with planning range shaping, the best-effort metric successfully selects better candidates. Meanwhile, planning range shaping contributes to better performance as well.

For the second aspect, we investigate if our proposed method can progressively navigate the robot to the final goal. 
Subsequently, we plan paths from the starting point or the last subgoal to the newly generated subgoal. The process terminates if the goal or the maximum number of generation trials of ten is reached. We replicate this process ten times for each problem.
Other than the random and best-effort selection mentioned in the first aspect, we add the goal-oriented selection mentioned in \ref{subsec:subgoal_select} for further comparison, marked as "G" in Table \ref{tab:exp2}.
Importantly, the success rate of reaching the \textit{final} goal is compared. 
Similar to the first aspect, we investigate how the planning time distribution of subgoal changes given different selection criteria. In addition, we compare the mean planning time for each subgoal and the complete path and employ path length in [rad] to depict the path quality. 
As the evaluation dataset, we select the planning problems whose planning time using RRTConnect is at least 0.05s in all runs. This leads to the result that the baseline method has a success rate of 100\%. Results are listed in Table \ref{tab:exp2}.

The results with the random selection indicate that a model relying solely on spatial information is not effective in either meeting time constraints or successfully guiding the robot to its goal configuration. Using the temporal information to select the candidates generally improves the performance in both aspects. 
The G-L-S variant generates sub-goals that can be planned with 0.030 seconds on average while the baseline method needs 1.15 seconds on average. Accumulating the planning time between all subgoals, our method achieved a result of 0.149 seconds on average, which is still way smaller than the baseline.
Compared to the AIT* solution, our approach exhibits a longer path length. An obvious reason is that we use the RRTConnect algorithm without path smoothing in the experiment, meaning that the path to the generated subgoals is not optimal. To examine the optimality of the subgoals, other metrics are needed.

% best-effort metric based on the temporal estimation improves the planning time and over 95\% of the selected candidates meet the time constraints. However, there are only minor improvements regarding the success rate. In all evaluated combinations, the combination of the goal-oriented metric and the best-effort metric shows the highest success rate while there is also a performance drop in the planning time of subgoals.

\subsection{Two robot arms experiment}
%We apply our approach to a bi-manual manipulation task setting. In a bi-manual manipulation task, while two robot arms are conducting the task in a decoupled fashion, they may have to traverse across each other. In this phase, generating the collision-free motion is very important. 
In this experiment, we apply the models trained by dataset mentioned in \ref{method:dataset} to a completely different scenario. We use two UR10e robot arms for goal reaching task, assuming the target configuration of the upstream manipulation task is given. These two robots are operating separately. %while knowing the state of each other. 
We apply our approach to one robot as motion planner, seeing the other robot as changing environment. This motion planner outputs a parameterized trajectory to the generated subgoals and finally completes the goal-reaching task without causing any collision. 

\begin{figure}
    \centering
    \includegraphics[width=0.49\textwidth]{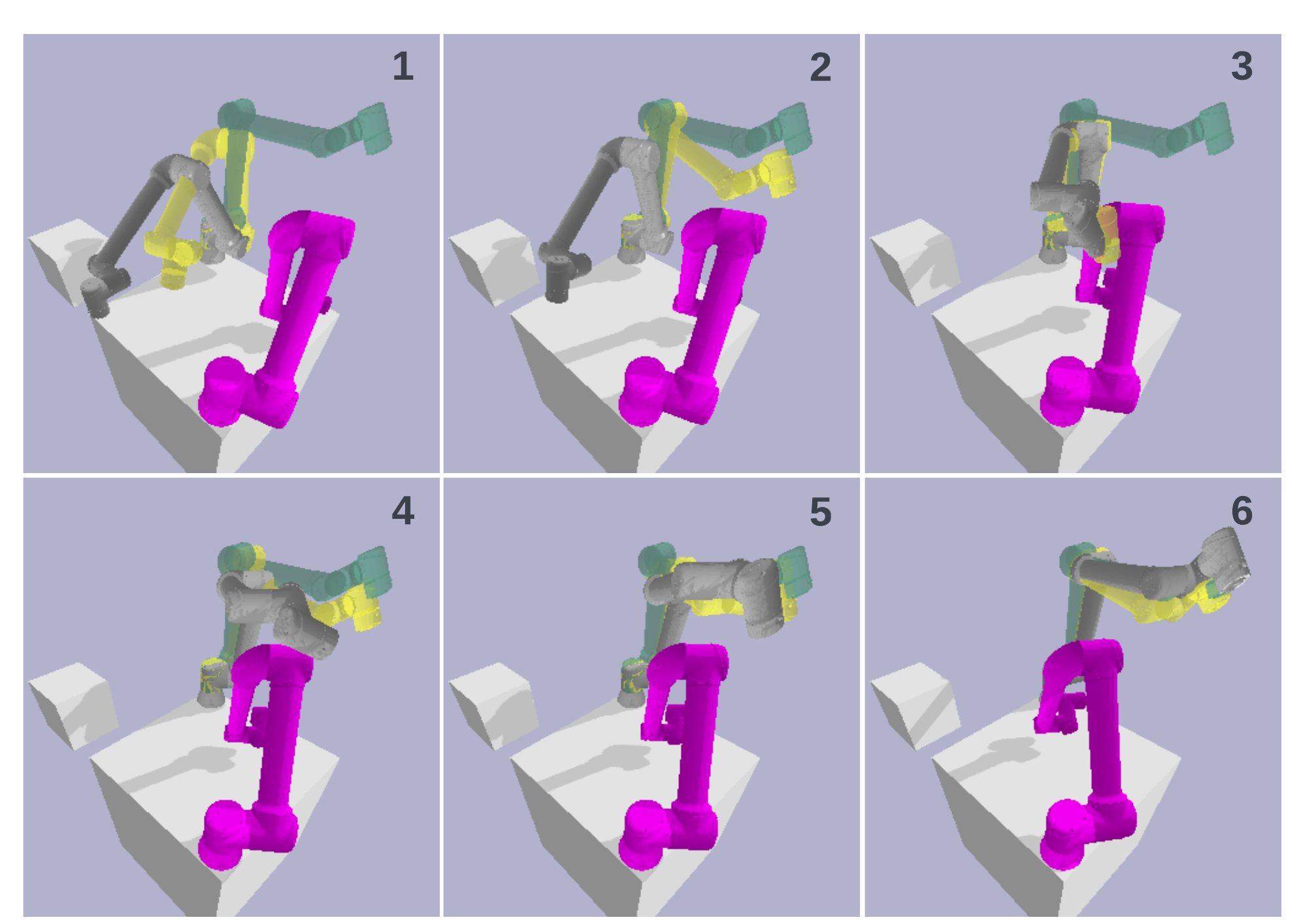}
    \caption{Two robot arms experiment. Gray: current G-robot statek. Green: final goal. Yellow: selected subgoals by PLS. Pink: P-robot as a moving object running a fixed trajectory.}
    \label{fig:two_robot}
\end{figure}

As shown in Fig \ref{fig:two_robot}, six snapshots are captured in chronological order. 
During the execution, the gray robot, we call it G-robot in the following, has to reach the final goal state presented in green. The yellow robot visualizes the selected subgoal. We use the log-normal time estimator and the goal-oriented metrics (G-L-S) for the selection. Meanwhile, the robot marked in pink, we call it P-robot in the following, follows a fixed trajectory to move from the right to the left.
% The snapshot with index 1 shows the initial state of the experiment. The first subgoal selected tends to guide the robot to move its end-effector to the base link. While the robot marked in pink is still far away, the model soon predicts a subgoal near the goal region in the snapshot with index 2. As the pink one continues to approach the middle of the table in 3, the generated subgoal leads the robot to move upwards to avoid colliding with the pink robot. Finally, the gray robot sweeps above the pink one, and the subgoal guides the robot to reach the goal region.  
The snapshot labeled with "1" presents the initial setup. The first subgoal chosen generally steers the G-robot's end-effector toward its base link, as it appears to be the shortest path to reach the goal when the P-robot is still at a considerable distance. In snapshot "2", 
the model predicts a subgoal close to the final goal region right away. As the P-robot approaches the center of the table in snapshot "3", %while the previous subgoal becomes timely infeasible, 
the newly generated subgoal directs the G-robot to elevate to avoid a collision. Finally, it maneuvers over the P-robot and reaches the final goal region.

\section{Conclusion}
In this work, we propose PLS that aims to generate subgoals that not only can lead a robot to its final goal but also can be planned from the current robot configuration within the desired period of time. We extend the existing generative models that use only spatial information by integrating temporal information to select proper subgoals from a set of candidates. Although experiments show that PLS can achieve shorter planning time for both subgoals and the accumulated path, we want to address two issues in the future, including the consideration of more temporal information including the movement of the obstacles and the previously generated subgoals, and the integration of time estimation closer into the generation.


%%%%%%%%%%%%%%%%%%%%%%%%%%%%%%%%%%%%%%%%%%%%%%%%%%%%%%%%%%%%%%%%%%%%%%%%%%%%%%%%

%%%%%%%%%%%%%%%%%%%%%%%%%%%%%%%%%%%%%%%%%%%%%%%%%%%%%%%%%%%%%%%%%%%%%%%%%%%%%%%%

%%%%%%%%%%%%%%%%%%%%%%%%%%%%%%%%%%%%%%%%%%%%%%%%%%%%%%%%%%%%%%%%%%%%%%%%%%%%%%%%
% \section*{ACKNOWLEDGMENT}

% ...

%%%%%%%%%%%%%%%%%%%%%%%%%%%%%%%%%%%%%%%%%%%%%%%%%%%%%%%%%%%%%%%%%%%%%%%%%%%%%%%%

\bibliographystyle{IEEEtran}
\bibliography{bib}
\end{document}